\documentclass[sensors,article,accept,pdftex,moreauthors]{Definitions/mdpi}

\firstpage{1} 
\pubvolume{23}
\issuenum{0}
\articlenumber{1234}
\pubyear{2023}
\copyrightyear{2023}
\externaleditor{Academic Editor: Emanuele Lindo Secco 
}
\datereceived{6 December 2022 %
}
\daterevised{6 January 2023 %
}
\dateaccepted{16 January 2023 %
}
\datepublished{20 January 2023 %
}

\hreflink{ https://doi.org/10.3390/s23031234} %

\usepackage{glossaries}
\usepackage{bm}
\usepackage{tensor}
\newacronym{hyq}{HyQ}{Hydraulically actuated Quadruped}

\newacronym{lf}{LF}{Left-Front}
\newacronym{rf}{RF}{Right-Front}
\newacronym{lh}{LH}{Left-Hind}
\newacronym{rh}{RH}{Right-Hind}

\newacronym{haa}{HAA}{Hip Adduction-Abduction}
\newacronym{hfe}{HFE}{Hip Flexion-Extension}
\newacronym{kfe}{KFE}{Knee Flexion-Extension}

\newacronym{imu}{IMU}{Inertial Measurement Unit}
\newacronym{dofs}{DoFs}{Degrees of Freedom}
\newacronym{rt}{RT}{Real Time}

\newacronym{com}{CoM}{Center of Mass}
\newacronym{cop}{CoP}{Center of Pressure}
\newacronym{zmp}{ZMP}{Zero Moment Point}
\newacronym{icp}{ICP}{Instantaneous Capture Point}
\newacronym{cp}{CP}{Capture Point}
\newacronym{cmp}{CMP}{Centroidal Moment Pivot}
\newacronym{grfs}{GRFs}{Ground Reaction Forces}

\newacronym{ls}{LS}{Least Square}

\newacronym{slip}{SLIP}{Spring Loaded Inverted Pendulum}
\newacronym{eom}{EoM}{Equation of Motions}
\newacronym{qp}{QP}{Quadratic Program}
\newacronym{sqp}{SQP}{Sequential Quadratic Programming}
\newacronym{mic}{MIC}{Mixed-Integer Convex}
\newacronym{cmaes}{CMA-ES}{Covariance Matrix Adaptation Evolution Strategy}
\newacronym{ara}{ARA*}{Anytime Repairing A*}
\newacronym{pca}{PCA}{Principal Component Analysis}
\newacronym{cpg}{CPG}{Central Pattern Generator}
\newacronym{wbc}{WBC}{Whole-Body Control}

\newacronym{mpc}{MPC}{Model Predictive Control}
\newacronym{nmpc}{NMPC}{Nonlinear Model Predictive Control}
\newacronym{ik}{IK}{Inverse Kinematic}
\newacronym{ocp}{OCP}{Optimal Control Problem}
\newacronym{nlp}{NLP}{Nonlinear Programming}
\newacronym{ltv}{LTV}{Linear Time Varying}

\newacronym{awbc}{c$^3$WBC}{Compliant Contact Consistent Whole-Body Control}
\newacronym{swbc}{sWBC}{Standard Whole-Body Control}
\newacronym{c3wbc}{c$^3$WBC}{Compliant Contact Consistent Whole-Body Control}
\newacronym{ste}{TCE}{Terrain Compliance Estimator}
\newacronym{c3}{\texttt{c}$^3$}{compliant contact consistent}

\newacronym{stance}{STANCE}{\textbf{S}oft \textbf{T}errain \textbf{A}daptation a\textbf{n}d \textbf{C}ompliance \textbf{E}stimation}

\newacronym{wbopt}{WBOpt}{Whole Body Optimization}

\newacronym{hc}{HC}{Hunt and Crossley's}
\newacronym{kv}{KV}{Kelvin-Voigt's}

\newacronym{wllsr}{WLLSR}{Weighted Linear Least Squared Regression}

\newacronym{mae}{MAE}{Mean Absolute Tracking Error}

\newacronym{ode}{ODE}{Open Dynamics Engine}

\newacronym{cmg}{CMG}{Control Moment Gyroscope}
\newacronym{ocs}{OCS}{Orientation Control System}
\newacronym{ddp}{DDP}{Differential Dynamic Programming}

\newcommand{\MF}[1]{\textcolor{black}{#1}}

\newcommand{\FR}[1]{\textcolor{black}{#1}}

\Title{Orientation Control System: Enhancing Aerial Maneuvers for Quadruped Robots}

\TitleCitation{Orientation Control System: Enhancing Aerial Maneuvers for Quadruped Robots}

\Author{Francesco %
 Roscia $^{1}$\orcidA{}, Andrea Cumerlotti $^{1,2}$, Andrea Del Prete $^{2}$\orcidB{}, Claudio Semini $^{1}$\orcidC{} and Michele Focchi $^{1,3,}$*\orcidD{}}

\AuthorNames{Francesco Roscia, Andrea Cumerlotti, Andrea Del Prete, Claudio Semini and Michele Focchi}

\AuthorCitation{Roscia, F.; Cumerlotti, A.; Del Prete, A.; Semini, C.; Focchi, M.}

\address{%
$^{1}$ \quad Dynamic Legged Systems (DLS) Lab, Istituto Italiano di Tecnologia (IIT), 16163 Genova, %
 Italy\\
$^{2}$ \quad Industrial Engineering Department (DII), University of Trento, 38123 Trento, Italy\\
$^{3}$ \quad Department of Information Engineering and Computer Science (DISI), University of Trento, 38123 Trento, Italy\\
}

\corres{Correspondence: michele.focchi@unitn.it}

\abstract{For legged robots, aerial motions are the only option to overpass obstacles that cannot be circumvented with standard locomotion gaits. In these cases, the robot must perform a leap to either jump onto the obstacle or fly over it. However, these movements represent a challenge, because, during the flight phase, the Center of Mass (CoM) cannot be controlled, and there is limited controllability over the orientation of the robot. This paper focuses on the latter issue and proposes an  Orientation Control System (OCS), consisting of two rotating and actuated masses (flywheels or reaction wheels), to gain control authority on the orientation of the robot. Due to the conservation of angular momentum, the rotational velocity if the robot can be adjusted to steer the robot's orientation, even when the robot has no contact with the ground. The axes of rotation of the flywheels are designed to be incident, leading to a compact orientation control system that is capable of controlling both roll and pitch angles, considering the different moments of inertia in the two directions. \MF{The concept was tested by means of simulations on the \mbox{robot Solo12}.}}

\makeglossaries

\begin{document}

\section{Introduction}\label{sec:introduction}
Legged robots are designed for traversing rough terrain.
Different types of gaits, such as trot \MF{\cite{barasuol2013reactive}} or crawl \MF{\cite{focchi2020heuristic}}, have been developed for quadrupedal robots. 
Thanks to progress over the last two decades, robots have become lighter and stronger, which has enabled them to perform with agile locomotion.
However, sometimes it is not possible for the robot to move around or over an obstacle with the gaits mentioned above, and jumps are required. 

When the robot is in the air, the CoM moves on a ballistic trajectory, and this is completely defined by the lift-off position and velocity. On the other hand, the base orientation can be changed so as to exploit the conservation of the system's angular momentum. This means that it is possible to control the base angular velocity by changing the inertia of the robot, e.g., changing the configuration of the joints. Nevertheless, the majority of quadrupeds are designed with light legs, resulting in limbs that have little influence on the total angular momentum.

Quadrupedal animals, like cats, can rearrange their tails and trunks to correct their orientations during a fall \cite{kane1969dynamical}.
Much work in the field of robotics has made use of an additional link, such as a tail, as in \cite{chu2019null,wenger2016frontal}.
This link rotates around an axis that does not pass through the robot's CoM. The distances between the axis of rotation and the CoM of both trunk and tail have a large effect on the total inertia, even with a small tail mass.
However, the placement of the additional link makes the resulting robot asymmetric. Moreover, due to its limited range of motion, a tail can be used only for a single jump, not for a repeated sequence \cite{johnson2012tail}.
\MF{To circumvent these drawbacks, in \cite{an2022design,an2020development,tang2022towards} the authors attached a morphable inertial tail with 3 Degrees of Freedom (DoFs) (pitch, yaw, and telescoping) on a monopod, a biped and a quadruped, respectively, to enhance the agility of locomotion and to improve safety in landing.}

It is possible to obtain a similar result by creating repetitive circular motions with the feet, like in \cite{hoffman2021exploiting}. \MF{Moving the feet outwards increases the robot's total inertia, so if a leg is extended during half of the motion and retracted in the other half, a net angular momentum results on the trunk, causing a rotation. In \cite{kurtz2022mini}} the authors proposed special heavy boots for Mini Cheetah and used a neural network to calculate online joint trajectories. However, this solution unnecessarily increased the inertia of the legs, complicating the locomotion problem, which could no longer rely on the massless leg assumption. 

Another option is to use a  Control Moment Gyroscope (CMG).
This consists of a wheel, spinning at a constant angular velocity, inside two or three actuated gimbals.
Tilting the wheel's axis of rotation generates gyroscopic torque.
This system is widespread in spacecraft reorientation \cite{yoon2002spacecraft}, but less frequently exploited in robot locomotion, either wheeled \cite{brown1996single} or legged \cite{mikhalkov2021gyrubot}.
The CMG presents interesting capabilities, but the presence of a pan-tilt unit to drive the gyroscope makes it impractical to mount it on a small, lig\mbox{htweight rob}ot.

Flywheels represent an additional option for controlling a robot's orientation.
Changing the angular velocity of a rotating mass attached to the trunk generates a torque that can reorient the robot. This device was first applied for spacecraft orientation \cite{oland2009reaction}, and only sporadically investigated in legged locomotion for controlling pitch orientation, both for bipeds \cite{Brown2016,xiong2020sequential} and quadrupeds \cite{kolvenbach2019towards,vasilopoulos2016quadruped}.
Compared with tails, flywheels do not have position limits, and, since their rotational axes pass through the CoM, angular momentum results in holonomic \cite{machairas2015quadruped}.
To get a fast response, it is necessary to have an abrupt change in the flywheel's angular velocity (angular acceleration).
Using a brake avoids the employment of a motor able to deliver higher torques \cite{gajamohan2012cubli}, 
keeping the system compact. The motor slowly accelerates the wheel to a certain speed to store angular momentum and, when a reorientation is required, the break stops its spin.
Since the effect of the break is unidirectional, a limitation of this approach lies in the fact that it is only possible to generate a rotation of the base in the opposite direction of the flywheel's angular velocity, making the approach unsuitable for applications where \textit{continuous} controllability is required. On the other hand, direct-drive controlled flywheels can create accelerations in both directions,  and can implement continuous control laws.

The benefits of this kind of OCS are multiple. 
\begin{enumerate}
\item Flywheels can correct orientation errors (e.g., due to disturbances or tracking inaccuracies in the angular momentum achieved at the lift-off) during flight    %
in a conti\mbox{nuous ma}nner. 
\item They can track a time-varying reference, e.g., the robot can land with a desired angular velocity (possibly zero) and orientation.
\item  Even in the presence of contact, the landing phase can be enhanced or dynamic gaits, such as a trot, stabilized, by significantly reducing trunk oscillations.  
\item  The presence of these additional joints, the only functions of which  are to control orientation, provides the possibility of relieving the efforts of the legs.
\end{enumerate}

In more complex scenarios, like for a somersault, limbs and OCS can operate in parallel to achieve a rotational angle larger than that achievable only with legs (e.g., due to to\mbox{rque limita}tion).

\subsection{Proposed Approach and Contribution}

\begin{figure}[h!]%
	\includegraphics[width=7cm,height=8cm,keepaspectratio]{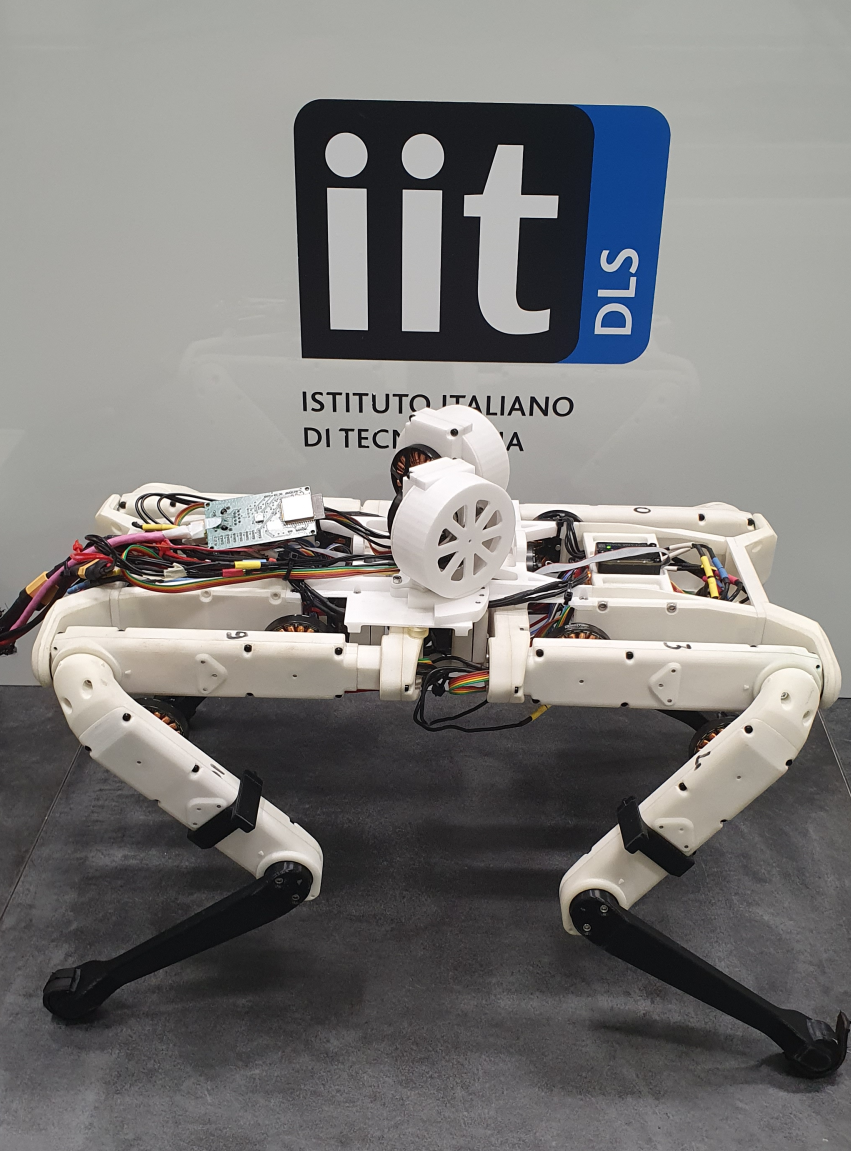}
	\caption{\small The proposed Orientation Control System for 
		for the $2.5 \ \mathrm{kg}$ quadruped robot Solo12 consists of two $0.1 \ \mathrm{kg}$ flywheels with incident rotation axes. Each wheel is located in a 3D-printed shell and mounted on the trunk of the body.}
	\label{fig:solo12flywheels}
\end{figure}

\MF{This work introduces} a compact OCS, based on two flywheels, mounted on the trunk of the lightweight quadruped Solo12 \cite{grimminger2020open}. \MF{The open-source hardware and software are reported in \cite{github_solo}. Figure \ref{fig:solo12flywheels} shows the OCS mounted on the trunk of the robot.} The contributions of the paper are the following:
\begin{itemize}
	\item the design of a novel OCS that enables effective control of the orientation of a legged robot during a jump, while keeping the design simple. 
	In particular, the axes of rotation of the flywheels are set to be incident, enabling \textit{continuous} %
 controllability in both directions of roll and pitch, while keeping the device compact.
	\item simulations  with the quadruped Solo12 were conducted and demonstrated the effectiveness of the proposed approach.
\end{itemize}

\subsection{Outline}
The remainder of this paper is organized as follows. In Section \ref{sec:background}, the law of conservation of the total angular momentum is described. In Section \ref{sec:ocs}, the design principles are presented, together with the strategy for simultaneously controlling the robot's roll and pitch orientations. Section \ref{sec:simulations} presents the simulation in different scenarios that demonstrate the capability of our OCS to reject disturbances and to track angular references when there is no contact with the ground, and to dampen base oscillation after touch-down. Conclusions and a possible evolution of the work are reported in Section \ref{sec:conclusions}.

\section{Background}
\label{sec:background}
The starting point for any OCS is the Euler's equation. For any mechanical system, the time derivative of the angular momentum, %
 $\bm{L}$ computed in respect to a reference point $O$, fixed in an inertial frame, equals the sum of the moments $\bm{M}_i$ applied to the system with respect to the same reference point:
\begin{equation}
\dot{\bm{L}} = \sum_i \bm{M}_i
\end{equation}
When %
 the result of the external moments applied to the system is zero, and the Euler's equation simplifies to:
 \begin{equation}
\dot{\bm{L}} = 0 \quad \Rightarrow \quad \bm{L}(t) = const,
\label{eq:cons_ang_mom}
\end{equation}
which is known as conservation of angular momentum.
Referring to legged robots, this condition occurs when the system is not in contact with the ground or with other objects, e.g., during a fall or the flight phase of a jump.
In this case, it is possible to change the angular velocity of the base link, thereby changing the joint positions and velocities, as a result of the \textit{non-holonomy} of the total angular momentum \cite{wieber2016modeling}. If the angular momentum of a certain body changes, then one of the others must change to maintain the total sum as constant.
\subsection*{Preliminary Analysis}
Most quadruped robots are designed in such a way that the largest amount of the mass is located in the main body. As a consequence, the contribution of the moments due to the acceleration of other bodies (i.e., leg links) is moderate. \\
As an example, consider the case of Solo12. Even if each limb accounts for about $13.4\%$ of the total mass, changes in the total inertia of the robot come with the motion of only the upper leg (thigh) and the lower leg (calf), since the displacement of distribution of the hip mass with the system's CoM is approximately constant. \MF{In view of this, each limb is responsible for only $7.5\%$ of the total mass in varying the total inertia of the robot}. Flywheels can be used to alleviate the lack of control authority.

\section{Orientation Control System}
\label{sec:ocs}
In this section, a procedure to select the inertia of the flywheels \MF{is presented, together with a strategy} to exploit them so as to simultaneously control the robot's roll and p\mbox{itch orien}tations.

\subsection{Bounds on the Inertia}
\begin{figure}[h!]%
	\includegraphics[width=.7\linewidth]{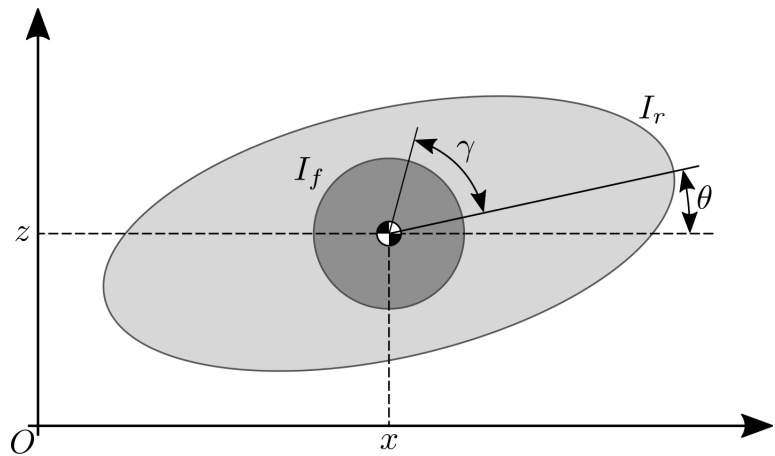}
	\caption{\small Schematic representation of the Elroy's Beanie model used for the preliminary analysis of the pitch motion.}
	\label{fig:elroy-beanie}
\end{figure}

\begin{figure}[h!]%
	\includegraphics[width=.8\linewidth]{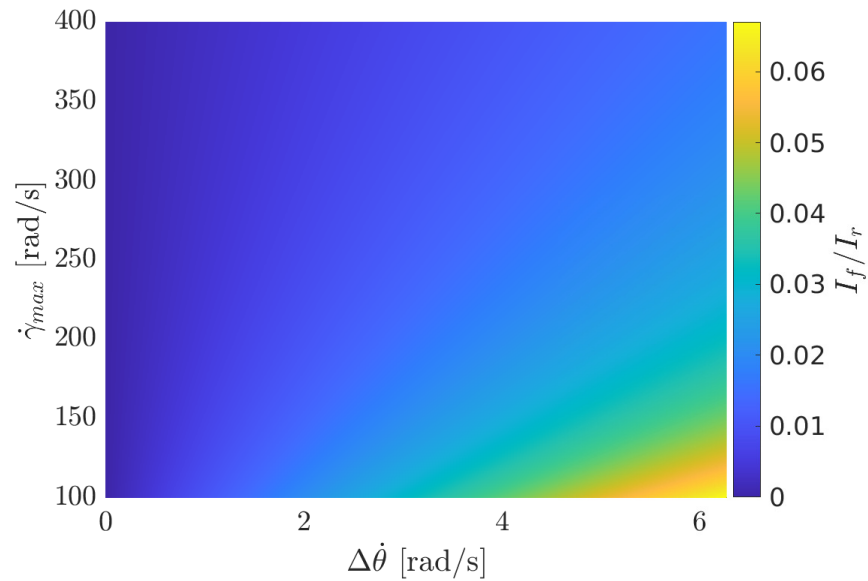}
	\caption{\small Minimum inertia of the OCS (normalized by the robot's inertia) necessary to achieve a base velocity variation $\Delta \dot{\theta}$, given the actuation bound $\dot \gamma^{max}$.}
	\label{fig:inertia_limits}
\end{figure}

Investigation for the selection of the flywheels' inertia can be performed with the Elroy's Beanie model, depicted in Figure \ref{fig:elroy-beanie}.
\noindent This consists of two rigid bodies connected through their CoMs with a revolute joint. One of the bodies represents the robot in its nominal configuration and the other models the OCS, here represented by a single wheel for the sake of simplicity. The aim was to examine the rotational dynamics of the system as a whole. To simplify the analysis, in the following \MF{only the effects on the pitch angle of the robot was considered}, keeping in mind that the same arguments also apply to the roll. 
Let us identify the moment of inertia of the robot in the nominal configuration as $I_r$ and the moment of inertia of the two flywheels as $I_f$, both referred to the system's CoM.
The angular momentum $L$ of this system can be written as:
\begin{equation}\label{eq:ang_mom_elroy}
L(t) = \left(I_r + I_{f}\right)\dot{\theta}(t) + I_{f}\dot{\gamma}(t)
\end{equation}
where $\dot{\theta}$ and $\dot{\gamma}$ are, respectively, the robot;s pitch rate and the angular speed of the wheel.
\MF{The robot is driven to reach} a desired pitch rate $\dot{\theta}_{des}$ by acting on the wheel speed.
Without loss of generality, \MF{one} can assume the flywheel is stationary with the robot at the instant at which the reorientation maneuver starts ($\dot{\gamma}_0 = 0$), having a system angular momentum of $L_0 = \left(I_r + I_{f}\right)\dot{\theta}_0$.
Under the condition of conservation of the angular momentum, this quantity is constant over time and it is possible to estimate the lower bound for $I_{f}$, given a desired pitch rate and the maximum velocity of the flywheels $\dot{\gamma}_{max}$:
\begin{equation}
I_{f} \geq I_r \dfrac{ \Delta \dot{\theta} }{\Delta \dot{\theta}+\dot{\gamma}_{max}}.
\end{equation}
where $\Delta \dot{\theta} =  \left\lvert \dot{\theta}_0-\dot{\theta}_{des} \right\rvert$ is the base velocity variation.
Figure \ref{fig:inertia_limits} reports the lower bound of $I_{f}$ given the desired base velocity variation and the maximum velocity of the actuator.

\vspace{-9pt}

\subsection{Flywheels' Axes of Rotation}
\label{subseq:flywheel_axes}

\begin{figure}[h!]%
	\includegraphics[width=7cm,height=8cm,keepaspectratio]{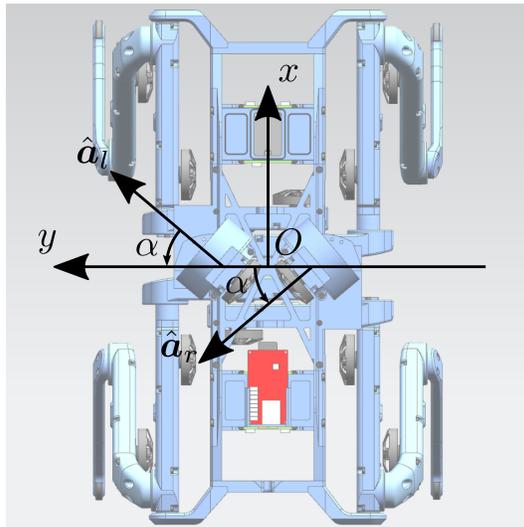}
	\caption{\MF{Top view of Solo12 with the proposed OCS. The $x-$ and $y-$axes embody the forward and left directions of the robot base, respectively. The figure also illustrates the unit vectors $\hat{\bm{a}}_l$ and $\hat{\bm{a}}_r$  that represent the axes of rotation of the left flywheel and of the right one and $\alpha$ is the angle of incidence, defined in Section \ref{subseq:flywheel_axes}.}}
	\label{fig:axes}
\end{figure}

\begin{figure}[h!]%
 \hspace{-16pt}
	\includegraphics[width=1\linewidth]{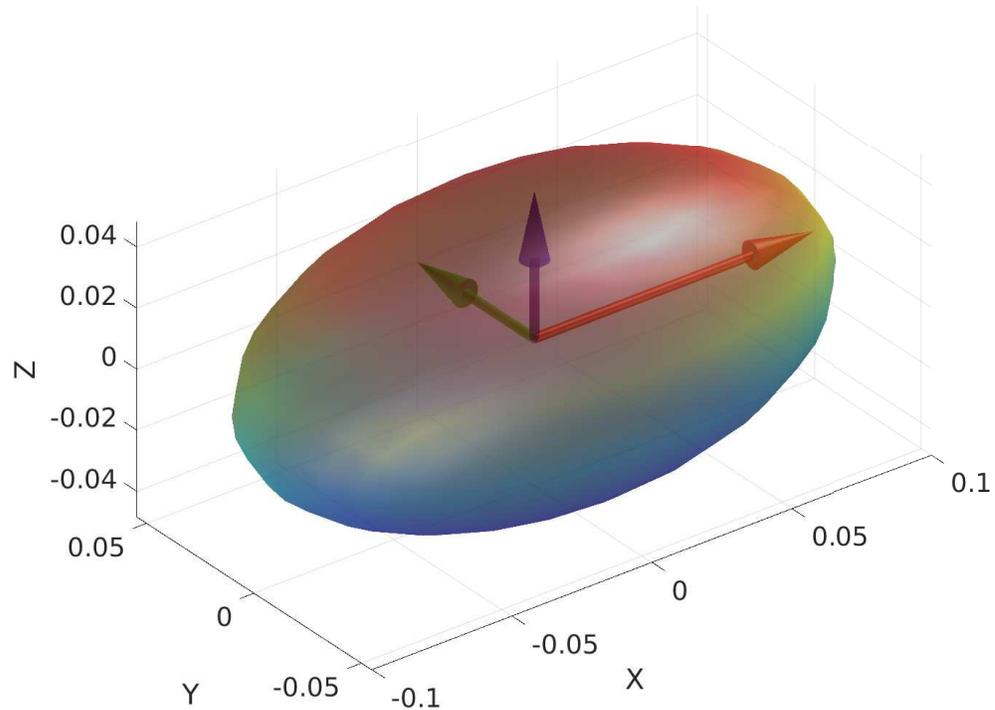}
	\vspace{-9pt}
	\caption{The %
 inertia tensor can be seen as an ellipsoid. Its principal axes are in the direction of the eigenvectors of the tensor and their \FR{lengths} depend on the eigenvalues.}
	\label{fig:ellipsoid}
\end{figure}

The orientation of the axis of rotation of each flywheel influences their contribution to the total angular momentum. To meet the specification of controlling both the robot roll and pitch, the axes of rotation of the left and right wheels, identified in the base reference frame with the unit vectors $\hat{\bm{a}}_l$ and $\hat{\bm{a}}_r$, are set to be incident, laying on a plane parallel to the $xy-$plane of the base reference frame (see Figure \ref{fig:axes}). 
To remove unnecessary complications, \MF{the authors} designed two identical modules, to be mounted on the trunk of Solo12, symmetrically, with respect to the robot's sagittal plane. \MF{Let} denote $\alpha \leq \pi/2$ with the non-negative incident angle between the wheels' axes of rotation and the robot's lateral direction. The matrix: 
\begin{equation}
\bm{C} = 
\left[ \begin{array}{cc}
\hat{\bm{a}}_l & \hat{\bm{a}}_r
\end{array} \right] = 
\left[ \begin{array}{cc}
\sin(\alpha) & -\sin(\alpha) \\
\cos(\alpha) & \cos(\alpha) \\
0 & 0
\end{array} \right]
\end{equation}
maps the flywheel moments into scalar torques $\bm{u} = \left[\begin{array}{cc} \tau_{fl} & \tau_{fr} \end{array} \right]^T$ about the flywheel axes $\hat{\bm{a}}_l$ and $\hat{\bm{a}}_r$ (expressed in the base frame).
As long as $\bm{C}$ is in full column rank, it is possible to control both roll and pitch angles. If $\alpha = 0$, the roll becomes uncontrollable through the OCS; otherwise, if $\alpha = \pi/2$, the pitch becomes uncontrollable. 
The angle is selected considering the ratio of the eigenvalues along the $x$ and $y$ directions of the ellipsoid of the robot's inertia(Figure \ref{fig:ellipsoid}):
\MF{\begin{equation}
\alpha^{*}=\tan^{-1}\left(\dfrac{I_{r, xx}}{I_{r, yy}}\right).
\end{equation}}

\vspace{-11pt}
\begin{figure}[h!]%
	\includegraphics[width=.7\linewidth]{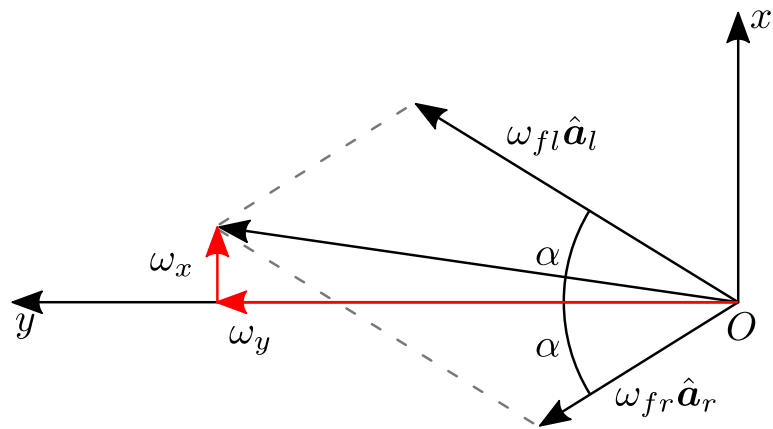}
	\caption{Having incident rotation axes, the OCS allows control of both the robot's roll and pitch. Notice that the roll angle is influenced by the difference of the angular speeds of the flywheels, $\omega_x = \left(\omega_{fl} - \omega_{fr}\right)\sin(\alpha)$. On the contrary, the pitch angle depends on the sum of the angular speeds $\omega_y = \left(\omega_{fl} +  \omega_{fr}\right)\cos(\alpha)$.}
	\label{fig:sum-diff}
\end{figure}

\noindent In the case of Solo12, this angle \FR{is} about $40^{\circ}$. With these considerations in mind, the angular momentum produced by the wheels, expressed in the robot base frame, was:
\begin{equation}
\begin{split}
\tensor[_b]{\bm{L}}{_f} &= \tensor[_b]{\bm{L}}{_{fl}} + \tensor[_b]{\bm{L}}{_{fr}} \\
&=I_{f, zz} \, \bm{\omega}_{fl} + I_{f, zz} \, \bm{\omega}_{fr}\\
&=I_{f, zz}
\left[\begin{array}{c}
\left(\omega_{fl} - \omega_{fr}\right)\sin(\alpha) \\ 
\left(\omega_{fl} +  \omega_{fr}\right)\cos(\alpha) \\
0
\end{array}\right]
\end{split}
\label{eq:bLf}
\end{equation}
in which $\bm{\omega}_{fl}= \omega_{fl} \, \hat{\bm{a}}_{l}$ and $\bm{\omega}_{fr}= \omega_{fr} \, \hat{\bm{a}}_{r}$ are the angular velocity vectors of the two wheels, and $\omega_{fl}$ and $\omega_{fr}$ are the angular speeds provided to each flywheel by its actuation system. The latter equation shows that the difference of the two angular speeds impacts on roll rotations, while their difference can be used to adjust the pitch, see Figure \ref{fig:sum-diff}. Using the definition of $\bm{C}$, \eqref{eq:bLf} is rewritten as:
\begin{equation}
\begin{split}
\tensor[_b]{\bm{L}}{_f} & = I_{f,zz}
\left[
\begin{array}{cc}
\sin(\alpha) & -\sin(\alpha) \\
\cos(\alpha) & \cos(\alpha) \\
0 & 0
\end{array}
\right]
\left[
\begin{array}{cc}
\omega_{fl} \\
\omega_{fr}
\end{array}
\right] \\
&= I_{f,zz} \bm{C} \bm{\omega}_f.
\end{split}
\end{equation}
\vspace{-6pt}

\subsection{Inertia Selection}
\begin{figure}[h!]%
	\includegraphics[width=7cm,height=8cm,keepaspectratio]{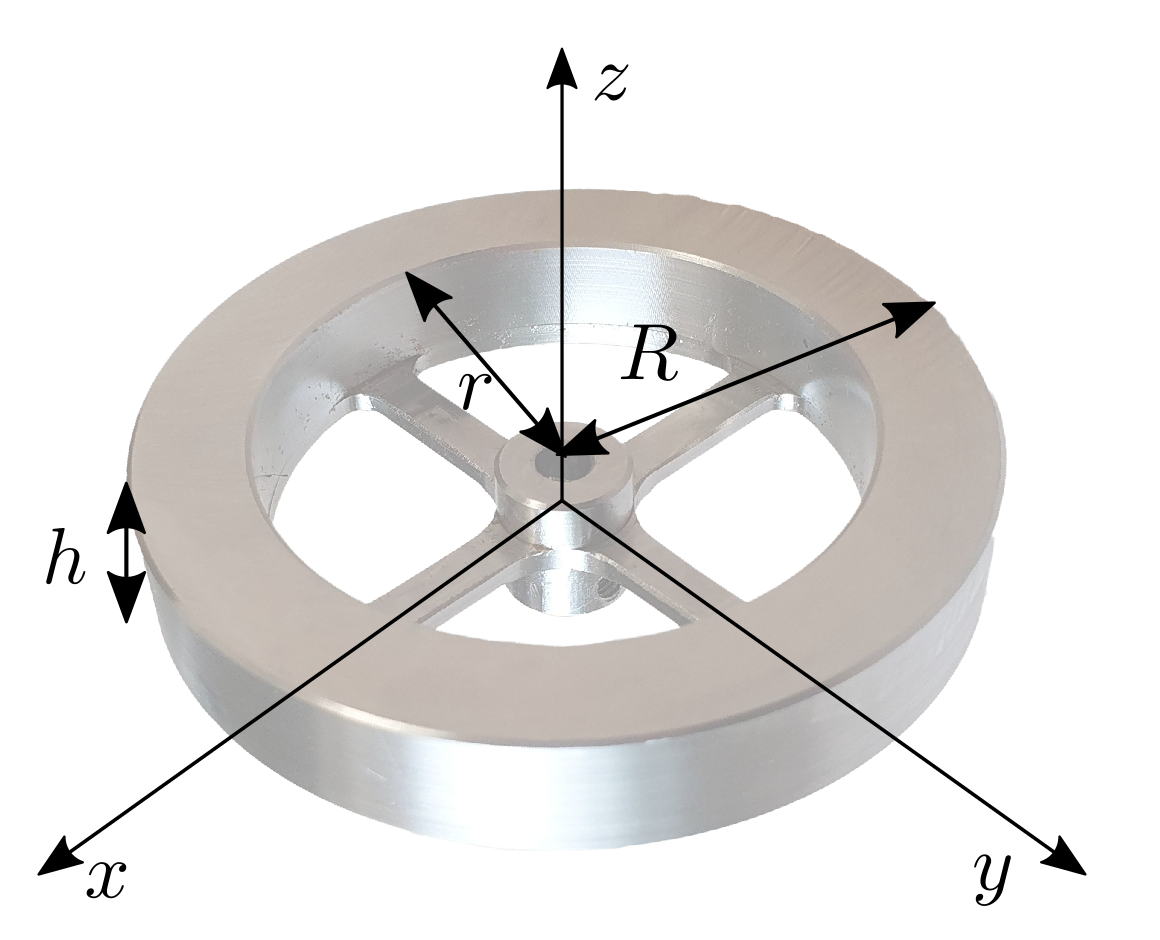}
	\caption{\small The final design of the flywheel, shown together with its principal axes of inertia. Here, $h$ is the wheel thickness, $r$ is the inner radius of the wheel and $R$ is the radius of the outer one.}
	\label{fig:flywheel}
\end{figure}

Once the desired inertia $I_f$ is selected, according to the 2D Elroy's Beanie model, it is used to realize the 3D OCS (Figure \ref{fig:flywheel}). 
\MF{Two identical flywheels were designed to have} the shape of hollow cylinders. This shape increases the inertia by locating the mass far away from the rotation axis. The inertia tensor expressed in its principal axes, \linebreak  $\bm{I}_f = \mathrm{diag}\left\{ I_{f, xx}, \, I_{f, yy}, \, I_{f, zz}  \right\}$, depends on the cylinders' inner and outer radii, $r$ and $R$, its height $h$ and material density $\rho$:
\begin{equation}
\begin{split}
I_{f, xx} = I_{f, yy} &= \frac{1}{12}\pi \rho h \left(3\left(R^4 - r^4\right) + h^2\left(R^2 - r^2\right)\right) \\
I_{f, zz} &= \frac{1}{2}\pi \rho h \left(R^4 - r^4\right)
\end{split}
\end{equation}
Notice that in the Elroy's Beanie model, there is a single body that models the complete OCS. The inertia introduced in the Elroy's Beanie model has to be split between the two flywheels: $I_{f, zz} = I_f/\left(2\cos(\alpha)\right)$. 
The parameters $R$ and $h$ can be set to have a compact OCS and $\rho$ depends on the chosen material, which was stainless steel in our case. The inner radius $r$ could be adjusted to obtain the desired inertia. Spokes with negligible mass connected the wheel to the motor shaft. All the parameters are reported in Table \ref{tab:dim_fw}, together with the selected inertia and mass of a single flywheel.

\vspace{-6pt}
\begin{table}[h!]%
	\caption{Sizes %
 and dynamic parameters of a single flywheel.}
	\label{tab:dim_fw}
	\newcolumntype{C}{>{\centering\arraybackslash}X}
	\begin{tabularx}{\textwidth}{CCC}
		\toprule
		\textbf{Parameter} & \textbf{Value} & \textbf{Unit}\\
		\midrule
		$r$ & $2.20 \times 10^{-2}$ & $\mathrm{m}$ \\
		$R$ & $3.00\times 10^{-2}$ & $\mathrm{m}$ \\
		$h$   & $1.02\times 10^{-2}$ & $\mathrm{m}$ \\
		$\rho$   & $7.86 \times 10^{+3}$ & $\mathrm{kg/m^3}$ \\
		$m$   & $1.02\times 10^{-1}$ & $\mathrm{kg}$ \\
		$I_{xx}, I_{yy}$   & $3.64 \times 10^{-5}$ & $\mathrm{kg \cdot m^2}$ \\
		$I_{zz}$   & $7.11 \times 10^{-5}$ & $ \mathrm{kg \cdot m^2}$\\
		\bottomrule	\end{tabularx}
\end{table}

\subsection{Flywheels' Control Law}
\label{subsec:control_law}
To derive a control law based on the robot's base orientation, we made use of \eqref{eq:cons_ang_mom}, expressing all the contributions to the time derivative of total angular momentum with respect to the base reference frame:
\begin{equation}
\tensor[_b]{\bm{I}}{_f} \, \tensor[_b]{\dot{\bm{\omega}}}{_f} + \tensor[_b]{\bm{I}}{_r} \tensor[_b]{\dot{\bm{\omega}}}{_r} + \tensor[_b]{\bm{\omega}}{_r} \times \left( \tensor[_b]{\bm{I}}{_r} \tensor[_b]{\bm{\omega}}{_r}\right) = 0
\end{equation}
From this expression, the moment on the base caused by the acceleration of the flywheels \MF{is defined}, and can be used as a feedback torque $\bm{\tau}_{fb}$:
\begin{equation}
\begin{split}
\bm{\tau}_{fb} &= \bm{I}_f \dot{\bm{\omega}}_f \\
&= - \tensor[_b]{\bm{I}}{_r} \dot{\bm{\omega}}_b - \tensor[_b]{\bm{\omega}}{_r} \times \left( \tensor[_b]{\bm{I}}{_r} \tensor[_b]{\bm{\omega}}{_r}\right)
\end{split}
\end{equation}
Proportional and derivative action $\bm{K}_p \bm{e} + \bm{K}_d \dot{\bm{e}_r}$ \MF{is used}, where $\bm{K}_p$ and $\bm{K}_d$ are diagonal and positive definite gain matrices for the error in attitude and angular velocity.
The orientation error $\bm{e} \in SO(3)$ requires that the algebra of the special rotational group is computed. To avoid singular configurations, orientations \MF{are represented} with quaternions.
The derivative error can be computed using $\dot{\bm{e}} = \tensor[_b]{\bm{\omega}}{_r^{des}} - \tensor[_b]{\bm{\omega}}{_r}$, in which $\tensor[_b]{\bm{\omega}}{_r^{des}}$ and $\tensor[_b]{\bm{\omega}}{_r}$ are, respectively, the desired and actual angular velocities of the base:
\begin{equation}
\bm{\tau}_{fb} = - \tensor[_b]{\bm{I}}{_r} \left(\bm{K}_p \bm{e} + \bm{K}_d \dot{\bm{e}}\right) - \tensor[_b]{\bm{\omega}}{_r} \times \left( \tensor[_b]{\bm{I}}{_r} \tensor[_b]{\bm{\omega}}{_r}\right).
\end{equation}
Projecting the moment $\bm{\tau}_{fb}$ onto the flywheel axes with $\bm{C}^T$, the control action $\bm{u}$ \MF{yields}
\begin{equation}
\label{eq:ctrl_law}
\bm{u} = \bm{C}^T \bm{\tau}_{fb}.
\end{equation}

\section{Results of Simulations}
\label{sec:simulations}
To validate our OCS, three simulations on different scenarios {are reported.~The authors} wanted to test the capability of the proposed approach in terms of the following: 
to reject a disturbance when the robot was in the flight phase of a jump, and to dampen trunk oscillations after touch-down, 
and to work in parallel with the joints of the legs to achieve a highly dynamic motion. 
All the simulations were performed using \MF{ROS \cite{quigley2009ros}, together with Gazebo \cite{koenig2004design}. The physics-related parameters are reported in Table \ref{tab:pys_param}.} References for the joints of the legs were computed off-line using Crocoddyl, an optimal control library for robots based on Differential Dynamic Programming (DDP) algorithms. It uses Pinocchio for fast computation of robot dynamics and the analytical derivatives. References $\bm{q}_{ref}$, $\dot{\bm{q}}_{ref}$ and $\bm{\tau}_{ref}$ for joint positions, velocities and torques were then executed on-line with a proportional-derivative joint controller:
\begin{equation}
\bm{\tau}_j = \bm{K}_{p,\, j} (\bm{q}_{ref} - \bm{q}) + \bm{K}_{d,\, j} (\dot{\bm{q}}_{ref} - \dot{\bm{q}}) + \bm{\tau}_{ref}
\end{equation}
The OCS was commanded to track base orientation references using the control law introduced in Section \ref{subsec:control_law} \MF{coded in Python}. (The %
video with all the simulations is available at web page %
\linksupplementary{s1}, see Supplementary Material)

\begin{table}[h!]%
	\caption{\MF{Physics-related parameters used to simulate the robot dynamics in Gazebo.}}
	\label{tab:pys_param}
	\newcolumntype{C}{>{\centering\arraybackslash}X}
	\begin{tabularx}{\textwidth}{CC}
		\toprule
		\textbf{Parameter} & \textbf{Value} \\
		\midrule
		Step size & $0.001 \ \mathrm{s}$\\
		Real time update rate & $250$ \\
		Physics engine & Open Dynamics Engine (ODE) \\
		Solver & Quick (Projected Gauss-Seidel method)\\
		Iterations & $50$\\
		Successive Over Relaxation parameter & $1.3$\\
		Rescaling Moment of Inertia  & no \\
		Friction model & Pyramid \\		
		\bottomrule	\end{tabularx}
\end{table}

\begin{figure}[h!]%
	\includegraphics[width=.7\linewidth]{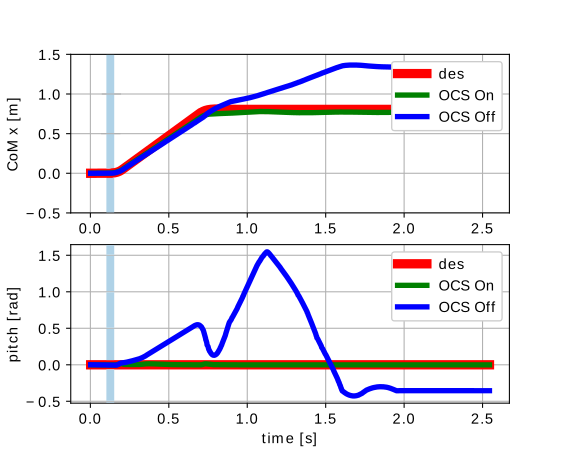}
	\caption{\small First Test: simulation results showing the $\mathrm{CoM}_x$ and pitch trajectory vs. time plots. A disturbance moment $\bm{\tau}_{dist}$ on the trunk could be compensated for only if the OCS was enabled. If it was disabled, the robot was unable to restore a safe configuration after touch-down and, eventually, fell. The light blue area represents the interval of time in which the disturbance was applied.}
	\label{fig:fw_jump}
\end{figure}

\textls[-15]{The necessity of having an OCS was revealed in the first test, as disturbance occurred when the robot had no contact with the ground. 
During the flight phase of a forward jump, $0.1~\mathrm{s}$ after the lift-off, an external disturbance moment was applied to the robot's base, deflecting the robot's orientation. This disturbance was set to $\bm{\tau}_{dist} = \left[\begin{array}{ccc}
0.2 & 0.8 & 0.0
\end{array}
\right]^T \ \mathrm{N \cdot m}$ and applied for $0.05 \ \mathrm{s}$. 
If the flywheels were not used, Solo12 fell after touch-down. Instead, when
using the flywheels %
it was possible to drive the robot to a safe configuration after landing (Figure \ref{fig:fw_jump}), without the need to implement a specific landing strategy, such as the one in \cite{jeon2021real}.
}

\begin{figure}[h!]%
	\includegraphics[width=.7\linewidth]{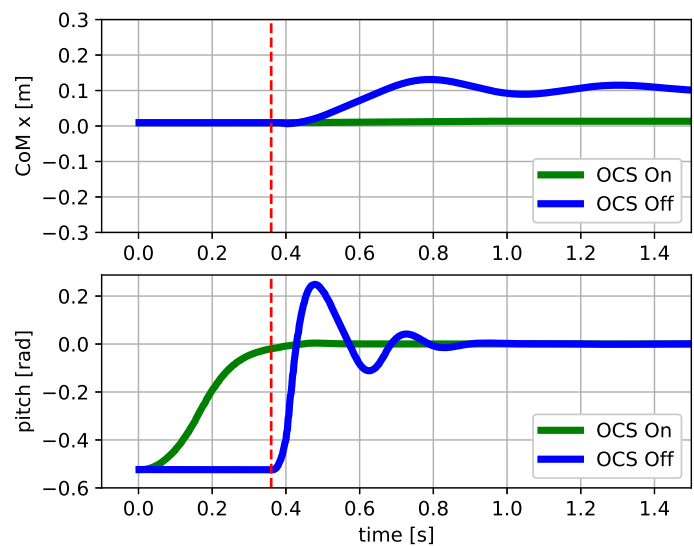}
	\caption{Second Test: simulation results showing the $\mathrm{CoM}_x$ and pitch trajectory vs. time plots. The OCS drove the robot's orientation during a fall so as to be horizontal. This allowed dampening of the base oscillations after touch-down (vertical dashed line), even without implementing a l\mbox{anding strat}egy.}
	\label{fig:fall_plot}
\end{figure}

\MF{The second test demonstrated} the ability of the OCS to reorient the base.
\MF{The robot fell} from a height of $1.0 \ \mathrm{m}$ with an initial pitch orientation of $30^\circ$ and zero base angular velocity.
\MF{If the flywheels were not actuated}, the robot touched the ground with the same initial orientation and the trunk oscillated. If the OCS was enabled, it drove the robot's base to be horizontal when it was still in the air and the oscillations after touch-down were drastically reduced, both in the pitch angle and forward direction (Figure \ref{fig:fall_plot}).

\begin{figure}[h!]%
	\includegraphics[width=\linewidth]{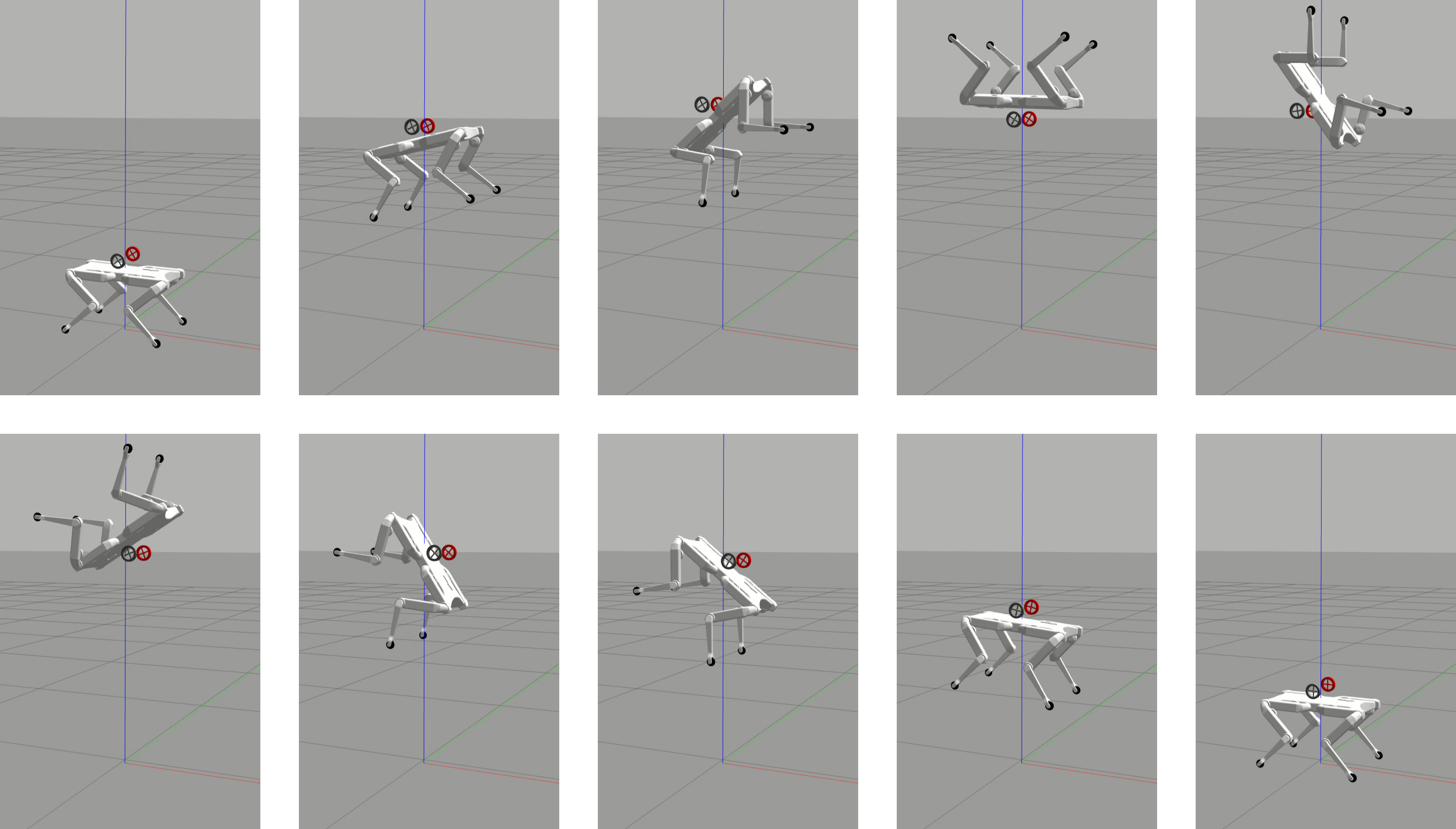}
	\caption{{Third Test: sequence of snapshots of Solo12 performing a back-flip in the Gazebo environment. The OCS alleviated the effort requested of the leg joints, that, in this case, could only be used to accomplish the vertical motion. The red, green and blue lines represent the axes of the inertial (world) reference frame.}}
	\label{fig:backflip}
\end{figure}

\MF{Finally, the ability of the flywheels to relieve the efforts of the leg joints to achieve a highly dynamic movement was shown with a back-flip} (Figure \ref{fig:backflip}). 
For this, we targeted a space application, carrying out a simulation with \MF{lunar gravity acceleration} ($1.62 \ \mathrm{m/s^2}$).
In this way, it was possible to obtain high jumps with a long flight phase, without having to select more powerful actuators.  
The leg joint trajectory computed off-line described a purely vertical jump of $1 \ \mathrm{m}$, having a flight phase that lasted $2 \ \mathrm{s}$.
Right after the lift-off, the flywheels started the reorientation task of performing a back-flip, by means of a spin of $360^\circ$ on the pitch.
For this maneuver, the value of the incident angle $\alpha$ was set to $0^\circ$, since no roll rotation was required.
Our hardware design allowed manual change of this value before performing the task.
This simulation demonstrated that the OCS alleviated the effort applied on the legs. Indeed, it would not have been possible to do a back-flip without the flywheels, because the legs were only used to achieve the linear motion. 
\vspace{-8pt}

\section{Conclusions}
\label{sec:conclusions}

\MF{This work presented} the design of a novel OCS, composed of two flywheels, that enable control of the trunk of a legged robot's platform, increasing the accuracy of aerial maneuvers during the under-actuated phases (i.e., flight phase), as well as enhancing stability when in contact by damping the oscillations. The novelty of the design, which involves flywheels attached with incident rotational axes on the trunk, is that it allows control of the orientations in both the roll and pitch directions, while keeping the device compact. The effect is optimized considering the inertial property of the mechanical structure. Several simulations were performed with the quadruped Solo12 that demonstrated the effectiveness of the proposed approach in the following: to reject disturbances during the flight phase, to stabilize the platform after touch-down, even in absence of a specific landing strategy, and to achieve a fast reorientation maneuver (a back-flip) in a reduced gravity environment. 

\MF{Possible applications of the presented OCS include, but are not limited to, efficient adjustment of the posture of quadruped robots walking or jumping on uneven terrains. As proved in the third simulation (back-flip), our approach improves the capabilities of quadrupeds in space application, enabling fast locomotion by means of leaps, ensuring a reactive control action on the robot's angular momentum. Furthermore, the method presented in Section \ref{sec:ocs} in designing the OCS does not depend on a specific platform, and, thus, it can be replicated for reorienting mechanical structures with different morphologies, e.g, monopods or bipeds. A construction worker's backpack could contain two flywheels with incident rotation axes, which, in the event of a fall from scaffolding, could be used to reorient a human body for impact on the ground with an upright posture, using the same controller proposed in this work.}

In future works, \MF{the authors} plan to demonstrate our concept performing experiments with a real platform. The control strategy for defining the base's desired angular velocity could be improved using  Nonlinear Model Predictive Control (NMPC), which takes into consideration future samples of the orientation reference.
\MF{This feature is expected to allow} the stabilization of the yaw to a desired value, so that, if it is not locally controllable, enhances the non-holonomy property of the angular momentum, i.e., doing a preliminary roll and pitch maneuver. 
A control strategy that may lead to the same result is dynamic feedback linearization, widely used for flight with quadricopters.

\vspace{6pt} 

 \supplementary{The %
 following supporting information can be downloaded at: \linksupplementary{s1},  Video S1: ocs\_video.mp4.%
}

\authorcontributions{Conceptualization, F.R. and M.F.; software, F.R., A.C. and M.F.; validation, F.R., A.C. and M.F.; formal analysis, F.R., A.C. and M.F.; investigation, F.R. and A.C.; resources, C.S.; data curation, F.R., A.C. and M.F.; writing---original draft preparation, F.R.; writing---review and editing, F.R., A.C., A.D.P., C.S. and M.F.; visualization, F.R. and A.C.; supervision, A.D.P. and M.F.; project administration, F.R. and M.F.; funding acquisition, M.F. and C.S. All authors have read and agreed to the published version of the manuscript.}

\funding{The publication was created with the co-financing of the European Union FSE-REACT-EU, PON Research and Innovation 2014-2020 DM1062/2021.}

\institutionalreview{Not applicable.}

\informedconsent{Not applicable.}

\dataavailability{Not applicable.} 

\acknowledgments{All the authors want to thank Roy Featherstone for interesting tips that started this work. }

\conflictsofinterest{The authors declare no conflict of interest.}

\abbreviations{Abbreviations}{
The following abbreviations are used in this manuscript:\\

\noindent 
\begin{tabular}{@{}ll}
DoF & Degree of Freedom\\
CoM & Center of Mass\\
OCS & Orientation Control System\\
CMG & Control Moment Gyroscope\\
\MF{DDP} & \MF{Differential Dynamic Programming}\\
NMPC & Nonlinear Model Predictive Control
\end{tabular}
}

\begin{adjustwidth}{-\extralength}{0cm}

\reftitle{References}

\PublishersNote{}
\end{adjustwidth}
\end{document}